\pdfoutput=1

\documentclass[11pt]{article}

\usepackage[]{acl}

\usepackage{times}
\usepackage{latexsym}

\usepackage[T1]{fontenc}

\usepackage[utf8]{inputenc}

\usepackage{microtype}

\usepackage{newfloat}
\usepackage{listings}
\usepackage{amssymb}
\usepackage{booktabs}
\usepackage{graphicx}

\title{IndicSUPERB: A Speech Processing Universal Performance Benchmark for Indian languages}
\author{
        Tahir Javed$^{1,2}$ \hspace{0.2cm} Kaushal Santosh Bhogale$^{1,2}$  \hspace{0.2cm} Abhigyan Raman$^{2}$ \hspace{0.2cm} \\
        \textbf{Anoop Kunchukuttan}$^{2,3}$ \hspace{0.2cm} \textbf{Pratyush Kumar}$^{2,3}$ \hspace{0.2cm}
         \textbf{Mitesh M. Khapra}$^{1,2}\thanks{ Corresponding author: miteshk@cse.iitm.ac.in}$ 
    \\ \\
    \textsuperscript{\rm 1}Indian Institute of Technology, Madras \\
    \textsuperscript{\rm 2}AI4Bharat \hspace{0.2cm} \textsuperscript{\rm 3}Microsoft \hspace{0.2cm}  \\
}

\newcommand{\dataset}{Kathbath}

\newcommand{\tick}{\makebox[0pt][l]{$\square$}\raisebox{.15ex}{\hspace{0.1em}$\checkmark$}}
\newcommand{\cross}{{x}}

\begin{document}

\maketitle

\begin{abstract}

A cornerstone in AI research has been the creation and adoption of standardized training and test datasets to earmark the progress of state-of-the-art models. A particularly successful example is the GLUE dataset for training and evaluating Natural Language Understanding (NLU) models for English. The large body of research around self-supervised BERT-based language models revolved around performance improvements on NLU tasks in GLUE. 
To evaluate language models in other languages, several language-specific GLUE datasets were created. The area of speech language understanding (SLU) has followed a similar trajectory. 
The success of large self-supervised models such as wav2vec2 enable creation of speech models with relatively easy to access unlabelled data. These models can then be evaluated on SLU tasks, such as the SUPERB benchmark. In this work, we extend this to Indic languages by releasing the IndicSUPERB benchmark. Specifically, we make the following three contributions.
(i) We collect \dataset~containing 1,684 hours of labelled speech data across 12 Indian languages from 1,218 contributors located in 203 districts in India. (ii) Using \dataset, we create benchmarks across 6 speech tasks: Automatic Speech Recognition, Speaker Verification, Speaker Identification (mono/multi), Language Identification, Query By Example, and Keyword Spotting for 12 languages. (iii) On the released benchmarks, we train and evaluate different self-supervised models alongside a commonly used baseline FBANK. We show that language-specific fine-tuned models are more accurate than baseline on most of the tasks, including a large gap of 76\% for the Language Identification task. However, for speaker identification, self-supervised models trained on large datasets demonstrate an advantage. We hope IndicSUPERB contributes to the progress of developing speech language understanding models for Indian languages. 

\end{abstract}

\section{Introduction}
Over the past few years, several works have shown that self-supervised learning (SSL) is very effective for natural language processing, vision and speech tasks \cite{DBLP:conf/naacl/DevlinCLT19, DBLP:conf/cvpr/NewellD20, DBLP:conf/interspeech/YangCCLLLLSCLHT21}. The key idea is to pre-train a large scale model on easily available unlabelled data and then adapt the same model for a wide variety of tasks by fine-tuning on smaller amounts of task specific data. Given the quantity and diversity of the pre-training data, such models learn to encode general purpose representations which can then be specialised for a wide variety of tasks. To evaluate the effectiveness of SSL, several diverse benchmarks such as GLUE \cite{DBLP:conf/emnlp/WangSMHLB18}, XTREME \cite{DBLP:journals/corr/abs-2003-11080}, SUPERB \cite{DBLP:conf/interspeech/YangCCLLLLSCLHT21}, \textit{etc.} have been created. Creating such benchmarks, while expensive and time consuming, has become very crucial for driving progress. In particular, languages for which such benchmarks do not exist often do not receive their fair share of representation in academic fora, thereby hampering progress \cite{joshi-etal-2020-state}.

\begin{figure}[!t]
   \centering
   \includegraphics[width=\linewidth]{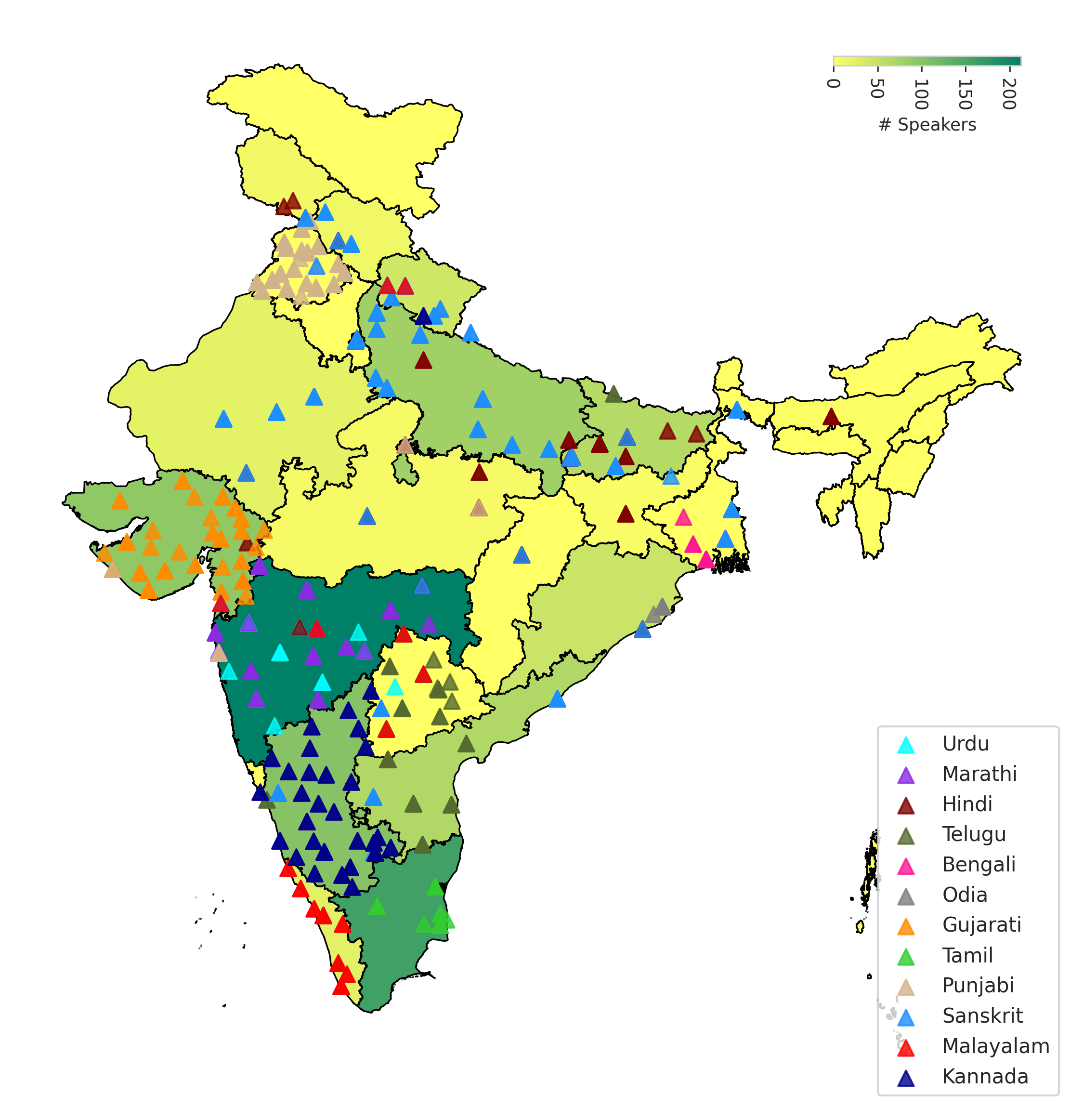}
   \caption{Distribution of Speakers across 22 Indian States}
   \label{fig:states}
\end{figure}

A case in point is that of languages from the Indian subcontinent. India has 22 constitutionally recognised languages with a collective speaker base of over 1 Billion speakers. On the one hand, there is an acute need for building speech and language understanding models for Indian languages, while on the other hand, there is an acute shortage of good benchmarks for training and evaluating such models. Recently released benchmarks such as IndicGLUE \cite{kakwani-etal-2020-indicnlpsuite}, and FLORES-200 \cite{nllb2022}  have alleviated the situation to a certain extent for natural language understanding and machine translation. 
However, for speech tasks, to the best of our knowledge, such a diverse benchmark is only available for English. In this work, we address this gap and build a Speech Language Understanding benchmark for 12 Indian languages covering 6 speech tasks: Automatic Speech Recognition, Speaker Verification, Speaker Identification (mono/multi), Language Identification, Query By Example, and Keyword Spotting.  

In this work, our first main contribution is to build a large dataset for automatic speech recognition containing read speech across 12 Indian languages from 1,218 speakers spanning 203 districts (see Figure \ref{fig:states}). This is a one of its kind open source effort resulting in a very large dataset containing 1,684 hours of labelled speech recognition data. The data was collected with the help of paid speakers using a \textit{maker-checker} flow. Every spoken utterance was verified by a human verifier to check that (i) there was no objectionable content in the audio, (ii) the recorded audio was faithful to the source text, and (iii) the recorded audio was clearly audible, \textit{i.e.},  it had appropriate volume and no background noise. For each of the 12 languages, we provided at least 100 hours of training data which helps in improving ASR models for these languages. Note that for 2 of the 12 languages no ASR training data is publicly available. Further, for 5 languages, our data helps in at least doubling the amount of training data in existing datasets.  We refer to this dataset as \dataset\footnote{Kathbath is a Kashmiri word relating to talks and conversations} and we hope that it will help in accelerating ASR research in Indian languages. 

Next, using \dataset~, we create a benchmark for four SLU tasks, \textit{viz.} automatic speech recognition, speaker verification, speaker identification, and language identification. For the latter three tasks, we use the speaker and language information associated with each utterance in \dataset~ to create training and evaluation sets with appropriate target labels. Further, to allow a robust evaluation of speech understanding and recognition models, we create multiple test conditions, \textit{viz.} (i) a test set containing speakers seen during training (ii) a test set containing novel speakers which are not seen during training (iii) a test set containing noisy data, \textit{i.e.}, data that was rejected during the verification stage (iv) a gender biased test set to evaluate the effect of having a higher number of female speakers in the training data. For speaker identification, we also create a language agnostic test set by pooling speakers from all the 12 languages. Apart from these four tasks, we also create a benchmark for query-by-example, wherein we ask speakers to utter a specific query and then collect all utterances containing this query. Lastly, we create a benchmark for the keyword spotting task, which involves classifying an utterance into a predefined set of keywords. We collectively refer to the above benchmark containing 6 tasks for 12 languages as IndicSUPERB.

We believe that using IndicSUPERB, it would be possible to answer several research questions which have previously not been addressed in the context of Indian languages. In this work itself, we address a few such questions: (i) How effective are SSL models as feature encoders for Indian languages? (ii) Are language-family specific SSL models better than those pre-trained on a larger set of languages? (iii) Are there specific layers in the network which are better suited for certain tasks? (iv) How does the performance generalise to unknown speakers? (v) Is there any gender-specific bias in the performance of these models? (vi) How robust are existing models to noise in the utterances? Based on our experiments, we observe that SSL models are good feature encoders with models pretrained using data from a diverse set of speakers being more suitable for speaker verification tasks. We also observe that while these models generalise well to unknown speakers in the test set and are reasonably robust to noise, they do get affected by specific biases (such as gender bias) in the training data. The main contributions of our work can thus be summarised as follows:
\begin{itemize}
    \item A speech recognition dataset, \dataset, containing a total of 1,684 hours of data from a diverse set of speakers across 12 Indian languages
    \item A robust evaluation benchmark, IndicSUPERB, for 6 speech language understanding tasks
    \item An analysis of the performance of existing SSL models using IndicSUPERB.
\end{itemize}

All the code, datasets and models developed as a part of this work have been made publicly available \footnote{https://github.com/AI4Bharat/indicSUPERB} and we hope that they will help in furthering research on speech technology for Indian languages.

\section{Related work}
In this section, we review existing datasets for SLU. As mentioned earlier, there are many datasets available for English, such as  Librispeech \cite{librispeech} and Common Voice for ASR, VoxCeleb1\cite{voxceleb1} and VoxCeleb2\cite{voxceleb2} for speaker identification/verification. More recently, SUPERB\cite{superb} and SUPERB-SG\cite{superbsg} have been released and contain benchmarks for various speech language understanding and synthesis tasks. In contrast, there are very few datasets for Indian languages as summarised in Table \ref{tab:related_work}. Most of these datasets such as MUCS\cite{diwan2021multilingual}, MSR\cite{srivastava18_sltu}, Gramvaani\footnote{https://sites.google.com/view/gramvaaniasrchallenge/dataset}, Crowd-source Speech Corpora (CSC) \cite{Kjartansson2018CrowdSourcedSC}, IISC-MILE Corpus \cite{mile_1}, Crowdsources Multispeaker Speech Dataset (CMSD) \cite{he-etal-2020-open}, Kashmiri Data Corpus (KDC)\footnote{https://www.openslr.org/122/}, Common Voice\footnote{https://commonvoice.mozilla.org/en/datasets}, IIIT-H Indic Speech Databases (ISD) \cite{prahallad2012iiit}, Hindi-Tamil ASR Challenge\footnote{https://sites.google.com/view/indian-language-asrchallenge/home} (HTAC), Vaksancayah (VAC)\cite{vacsa}, IIIT-H Telugu Corpus\footnote{https://asr.iiit.ac.in/cstd/}, IITB Marathi Corpus (IMC) \cite{marathidata} and SMC Malayalam Corpus \footnote{https://blog.smc.org.in/malayalam-speech-corpus/} contain only ASR data. Further, most of them support very few languages. 
One interesting effort worth mentioning is the LDCIL repository, which contains datasets for 22 Indian languages with all the meta information about speakers, language, gender and hence can be used to create a complete benchmark covering all the tasks. However, this data is not freely available and is shared selectively after a thorough vetting process to ensure non-commercial usage. To the best of our knowledge, ours is the first effort to create a robust benchmark for a diverse set of SLU tasks across 12 Indian languages.

\begingroup
\renewcommand{\arraystretch}{0.7} 
\begin{table*}[ht]
\centering
\small
\begin{tabular}{@{}lcccccccccccccc@{}}
\toprule
Tasks &\textbf{ours} & \textbf{mucs} & \textbf{msr} & \textbf{csc} & \textbf{iisc} & \textbf{cmsd} & \textbf{kdc} & \textbf{cv} & \textbf{isd} & \textbf{htac} & \textbf{vac} & \textbf{iiith-t} & \textbf{imc} & \textbf{smc}\\
\midrule
ASR & \tick & \tick & \tick & \tick & \tick & \tick & \tick & \tick & \tick & \tick & \tick& \tick & \tick& \tick\\
SID & \tick & \cross & \cross & \cross & \cross & \cross & \cross & \cross & \cross & \cross& \cross& \cross& \cross& \cross \\
LID & \tick & \cross & \cross & \cross & \cross & \cross & \cross  & \cross & \cross & \cross& \cross& \cross & \cross& \cross\\
ASV & \tick & \cross & \cross & \cross & \cross & \cross& \cross & \cross & \cross & \cross& \cross& \cross & \cross& \cross\\
QbE & \tick & \cross & \cross & \cross & \cross& \cross & \cross & \cross & \cross & \cross& \cross& \cross& \cross& \cross\\
KS & \tick & \cross & \cross & \cross & \cross& \cross & \cross & \cross & \cross & \cross& \cross& \cross& \cross& \cross\\
\midrule
\#Lang & 12 & 6 & 3 & 3 & 2 & 6 & 1 & 6 &7&2& 1 & 1& 1 &1 \\
\#ATH & 140 & 67 & 40 & 206 & 214 & 3 & 3 & 60 & 2 &141&56 &2000&109& 1 \\

\bottomrule
\end{tabular}
\caption{Table describing the availability of tasks across different publicly available benchmarks for Indian Languages. (\#Lang = Number of Indian languages supported, \#ATH = average training hours per language for ASR task)}
\label{tab:related_work}
\end{table*}
\endgroup

\section{Data Collection Methodology}
The first step towards creating IndicSUPERB was to collect voice samples and their transcriptions from a diverse set of speakers for multiple Indian languages. Once such voice samples were collected, we could use the meta-data associated with them (such as speaker/language information) to create data for different tasks as mentioned earlier. We now describe the various choices that we made for collecting these voice samples which form the core of our dataset.

    \noindent\textbf{Languages.} Ideally, we would have liked to create IndicSUPERB for all the 22 constitutionally recognised languages in India, but based on budget restrictions we had to choose 12 languages. We decided to select these languages such that there is a fair representation of languages from North, South, East and West of India. The specific languages that we selected were Kannada, Malayalam, Tamil and Telugu from South India, Gujarati and Marathi from West India, Bengali and Odia from East India,  Hindi and Punjabi from North India, and Sanskrit and Urdu which are spoken in different parts of the country. 
    
    \noindent \textbf{Type of data.} While collecting speech recognition data, there are three main options that one could consider. The first option is to collect read-speech wherein participants are shown a piece of text and then asked to speak it out. The second option is to ask participants to speak on a topic without a script and then transcribe the data. The third option is to curate existing audio content from the web (such as YouTube) and transcribe it. In line with similar efforts for low resource languages (such as the Mozilla Common Voice project), we decided to collect read speech data which is cheaper. The other advantage is that it is easier for speakers to participate as opposed to collecting unscripted extempore data where many participants may not be able to speak more than a few words. Further, unlike YouTube data where there are issues in finding the meta-data associated with the speaker (age, geography,\textit{ etc.}), here the participants can explicitly provide such information. Such meta-data is important for measuring the diversity of the data as well as for creating data for downstream tasks, such as speaker identification.
    
    \noindent \textbf{Selection of text.} We used IndicCorp \cite{kakwani-etal-2020-indicnlpsuite} for collecting sentences which can then be read by speakers. IndicCorp is the largest publicly available collection of monolingual corpora for Indian languages collected from a diverse set of India-specific sources on the web (such as news articles, government websites, \textit{etc}). We collected $\sim$100K sentences for each of the 12 languages from IndicCorp while restricting the character set to alphanumeric only and the sentence length to 8-15 words. This was done to ensure that the sentences were clean and small enough to be spoken fluently by the annotators. The size of the vocabulary derived from these sentences for each of the 12 languages is summarized in Table \ref{tab:data_sum_all}.

    \noindent \textbf{Tool for data collection and verification.}  
    Our data collection started during the middle of the COVID-19 pandemic making it difficult for participants to physically assemble at one location to record voice data. We thus needed a tool which allowed us to distribute and track work remotely to a large number of participants who were geographically spread out. To do so, we used Microsoft's open source crowdsourcing platform called Karya which was already used by other teams in the past to collect voice data \cite{abraham-etal-2020-crowdsourcing}. Karya is available as an android application which can run offline and sync with the backend server at periodic intervals to store the collected data. This makes it ideal for use in Indian scenarios where continuous connectivity is sometimes an issue. For every language, we uploaded 100K sentences into Karya which could be accessed in batches of 100. The distribution of tasks was managed with the help of access codes. In particular, Karya generated an access code for each batch of 100 sentences. These access codes were shared with participants who could then log in to the tool and work on the corresponding tasks. Once a task is completed, Karya initiates a verification workflow wherein the collected data is verified by a human. 
    
    \noindent \textbf{Ensuring quality control.} To ensure that the recorded data is of high quality, it is essential that (i) the text data does not contain any offensive/objectionable content (ii) the audio samples are accurate, \textit{i.e.}, they contain the same content as in the text data (iii) the audio samples are clearly audible with no background ambient noise. To ensure this, we took the following steps. First, during the recording stage, the participants were strictly instructed to skip any sentences which contained offensive/objectionable content. In addition, each recorded sample was also passed through a verification stage, where the verifiers were again asked to discard any recordings which contained objectionable content even if they were accurate and audible. Apart from eliminating all objectionable content, the verifiers were asked to score each recorded sample on 3 parameters: accuracy, volume, and quality. Each score ranges from 0 to 2. A recorded sample would be given a score of 2 on accuracy only if its contents exactly matched the source sentence. Similarly, a score of 2 for volume would be given only if the recorded sample was clearly audible. Lastly, a score of 2 on quality would be given only if the recorded sample had no background noise. We accepted only those samples which had a score of 2 for all the 3 parameters.
    
    During the initial phases of data collection, we observed that nearly 60\% of the data was getting discarded during the verification stage. On further analysis, we found that the main reasons for rejection were (i) inaccurate reading of the source text (ii) presence of background noise. To avoid inaccurate recordings, we requested the annotators to practice by reading the sentence aloud a couple of times before actually recording it. We also asked them to skip a sentence if they were unsure about the pronunciation of certain words in the sentences or found certain words to be difficult to pronounce. To avoid background noise, we requested participants to record sentences during the night when noise from vehicles, household chatter, \textit{etc.} was minimal. With these instructions, we found a significant reduction in the rejection rate. Table \ref{tab:data_sum_all} summarises the total amount of clean and noisy data that was collected as a part of the process. 
    
    \begin{figure}[h]
   \centering
   \includegraphics[width=\linewidth]{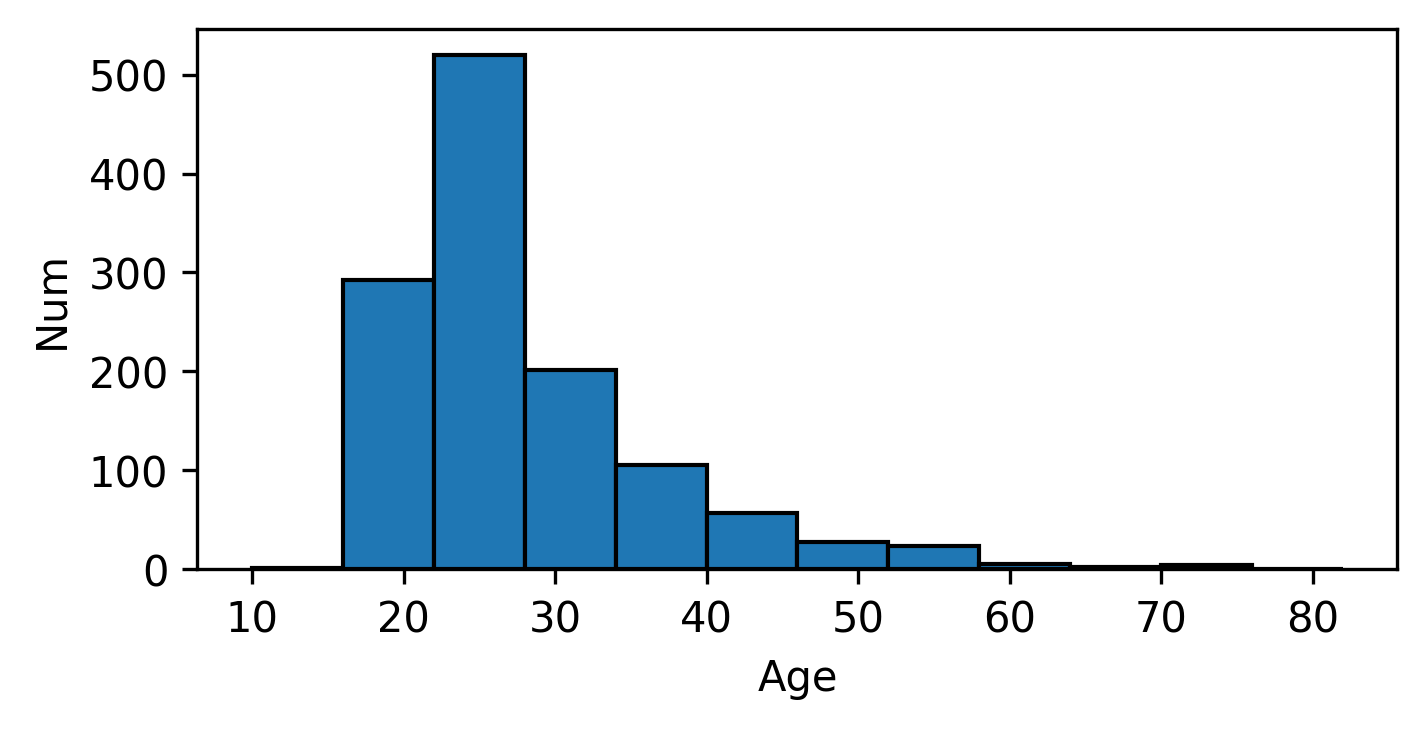}
   \caption{Age-wise distribution of speakers in Kathbath}
   \label{fig:barplot_age}
\end{figure}

    \noindent \textbf{Ensuring diversity.} To ensure diversity in the collected data we ensured that the participants came from different districts, different age groups and genders. In total, we collected data from 1,218 speakers spanning 203
    districts across 22 states in India. While for most languages, we had a nearly 50-50 representation for males and females, for some languages, we ensured that the collected data had a much higher proportion of female speakers (as female participants are often poorly represented in many AI datasets). This would allow a more systematic study of the effect of bias in training data. Table \ref{tab:data_sum_all} shows the total number of male and female speakers in each language. Figure \ref{fig:states} shows the different states from which the data was collected. Figure \ref{fig:barplot_age} shows the age-wise distribution of speakers in the dataset. 

    \begingroup
    \setlength{\tabcolsep}{2pt} 
    \renewcommand{\arraystretch}{0.7} 
    \begin{table*}[ht]
    \centering
    
    \small
    \begin{tabular}{@{}lcccccccccccc@{}}
    \toprule
    \textbf{} &\textbf{bn} &\textbf{gu} &\textbf{hi} &\textbf{kn} &\textbf{ml} &\textbf{mr} &\textbf{or} &\textbf{pa} &\textbf{sa} &\textbf{ta} &\textbf{te} &\textbf{ur} \\
    \midrule
    Clean data duration (hrs) &115.8 &129.3 &150.2 &165.8 &147.3 &185.2 &111.6 &136.9 &115.5 &185.1 &154.9 &86.7 \\
    Noisy data duration (hrs) &38.8 &65.1 &47.7 &64.5 &15.7 &86.5 &27.6 &36.5 &76.3 &95.8 &74.8 &77.0 \\
    \midrule
    \# Male speakers  & 18 & 44 & 58 & 53 & 12 & 82 & 10 & 65 & 95 & 116 & 53 &36 \\
    \# Female speakers & 28	&35	&63	&26	&20	&61	&32	&77	&110 &42& 51&31\\
    \midrule
    \# Unique words  & 6k & 109k  & 54k & 181k & 268k & 132k & 94k & 56k & 298k & 171k & 147k & 44k \\
    \bottomrule
    \end{tabular}
    \caption{Table summarizing number of hours of data, number of speakers and vocabulary size in Kathbath.}
    \label{tab:data_sum_all}
    \end{table*}
    \endgroup
    \noindent \textbf{Consent of participants.} For every language, we had a local coordinator hired through a data collection agency who clearly explained the nature of the activity to the participant. The local coordinator ensured that each participant understood that his/her voice samples were being collected for research purposes and that the data will be publicly released and used for building AI models. The terms and conditions and the nature of the activity were also clearly shown when the participants first logged into Karya. The app would proceed to recording only after the participants read the instructions and provided their consent. 
    
    \noindent \textbf{Cost of data collection.} The cost of data collection was INR 1500 per hour (approximately, 25 USD per hour). Each participant was paid between INR 500 - 1000 for 1 hour of recording (approximately, 6 to 12 USD per hour).

\section{IndicSUPERB}
The above data collection process resulted in a total of 1,684 hours of read speech data across 12 languages with a total of 0.9M utterances. Note that each utterance in the data is unique, \textit{i.e.}, a given sentence is only spoken by one speaker and never repeated again in the data.
We now explain the procedure for creating the IndicSUPERB benchmark from this data. 

\subsection{Train, validation, test splits}
Our goal was to provide training as well as evaluation data for various SLU tasks. Further, we wanted that the benchmark should support different conditions, \textit{e.g.}, (i) evaluation for speakers existing in the training data (ii) evaluation for speakers not existing in the training data (iii) evaluation on noisy data. To enable this, we divided the data into the training set and multiple validation and test sets as explained below. 

\noindent\textbf{Test-Unknown.} For each of the 12 languages, we first set aside audio recordings corresponding to 10 male and 10 female speakers contributing a total of 3 hours (1.5 hours each for male and female). We then removed all the instances of these speakers from the rest of the data. This ensured that these speakers are only seen at test time and are unseen during training and validation.

\noindent\textbf{Test-Known.} For creating a test set containing known speakers, for each of the 12 languages, we again take 10 male and 10 female speakers and sample audio recordings for them containing a total of 5 hours (2.5 hours each for male and female). Unlike before, we do not remove other instances of these speakers from the rest of the data. Hence, some data for these speakers is also seen during training.

\noindent\textbf{Validation.} We create the validation data using the same procedure as that used for creating the test-known set. Note that some of the speakers in the test-known are also present in the validation data. 

\noindent\textbf{Training data.} We use all the remaining data as training data. We reiterate that some of the speakers in the validation and test-known splits are also present in the training data but none of the speakers in the test-unknown split is present in the training data. Further, since each utterance in our data is unique, no sentence in the training data is present in any of the other splits. 

Next, we create some noisy test and validation sets which would help in evaluating the robustness of speech models. For this, we first consider the data that was rejected during the verification stage and extract all recordings which had a rating of at least 1 for all the 3 parameters (accuracy, volume and quality). We refer to this as the noisy set and create the following splits from it.

\noindent\textbf{Noisy-Test-Unknown.} A total of 3 hours of noisy data from 10 male and 10 female speakers which are not seen in the training data described above.

\noindent\textbf{Noisy-Test-Known.} A total of 5 hours of noisy data from 10 male and 10 female speakers which are also seen in the training data described above. 


Table \ref{tab:language_wise_distribution} summarises the above splits for all the 12 languages.

\begingroup
\setlength{\tabcolsep}{6pt} 
\renewcommand{\arraystretch}{0.8} 
\begin{table*}[ht]
\centering
\small
\begin{tabular}{@{}lcccccccccccc@{}}
\toprule
\textbf{} &\textbf{bn} &\textbf{gu} &\textbf{hi} &\textbf{kn} &\textbf{ml} &\textbf{mr} &\textbf{or} &\textbf{pa} &\textbf{sa} &\textbf{ta} &\textbf{te} &\textbf{ur} \\
\midrule
\multicolumn{13}{@{}l}{\textbf{Train}}  \\
\midrule
\# M &10 &34 &48 &43 &7 &72 &4 &55 &85 &106 &43 &26 \\
\# F &18 &25 &53 &16 &10 &51 &22 &67 &100 &32 &41 &21 \\
MD(h) &24 &28 &42 &56 &46 &67 &49 &20 &31 &80 &52 &22 \\
FD(h)&79 &88 &95 &97 &89 &105 &49 &104 &72 &92 &90 &52 \\
\midrule
\multicolumn{13}{@{}l}{\textbf{Valid}}  \\
\midrule
\# M &9 &10 &10 &10 &6 &10 &4 &10 &10 &10 &10 &10 \\
\# F &10 &10 &10 &10 &10 &10 &10 &10 &10 &10 &10 &10 \\
MD(h) &2.5 &2.5 &2.5 &2.5 &2.5 &2.5 &2.5 &2.5 &2.5 &2.5 &2.5 &2.5 \\
FD(h) &2.5 &2.5 &2.5 &2.5 &2.5 &2.5 &2.5 &2.5 &2.5 &2.5 &2.5 &2.5 \\
\midrule
\multicolumn{13}{@{}l}{\textbf{Test Known}}  \\
\midrule
\# M &10 &10 &10 &10 &7 &10 &4 &10 &10 &10 &10 &10 \\
\# F&10 &10 &10 &10 &10 &10 &10 &10 &10 &10 &10 &10 \\
MD(h)&2.5 &2.5 &2.5 &2.5 &2.5 &2.5 &2.5 &2.5 &2.5 &2.5 &2.5 &2.5 \\
FD(h) &2.5 &2.5 &2.5 &2.5 &2.5 &2.5 &2.5 &2.5 &2.5 &2.5 &2.5 &2.5 \\
\midrule
\multicolumn{13}{@{}l}{\textbf{Test Unknown}}  \\
\midrule
\# M &8 &10 &10 &10 &5 &10 &6 &10 &10 &10 &10 &10 \\
\# F &10 &10 &10 &10 &10 &10 &10 &10 &10 &10 &10 &10 \\
MD(h) &1.7 &1.5 &1.5 &1.5 &1.5 &1.6 &1.5 &1.5 &1.5 &1.5 &1.4 &1.5 \\
FD(h) &1.5 &1.5 &1.5 &1.5 &1.5 &1.5 &1.5 &1.5 &1.5 &1.7 &1.7 &1.6 \\
\midrule
\multicolumn{13}{@{}l}{\textbf{Noisy Test Known}}  \\
\midrule
\# M &10 &10 &10 &10 &7 &10 &4 &10 &10 &10 &10 &10 \\
\# F &10 &10 &10 &10 &10 &10 &10 &10 &10 &10 &10 &10 \\
MD(h) &2.5 &2.5 &2.5 &2.5 &2.5 &2.5 &2.5 &2.5 &2.5 &2.5 &2.5 &2.5 \\
FD(h) &2.5 &2.5 &2.5 &2.5 &2.5 &2.5 &2.5 &2.5 &2.5 &2.5 &2.5 &2.5 \\
\midrule
\multicolumn{13}{@{}l}{\textbf{Noisy Test Unknown}}  \\
\midrule
\# M &8 &10 &10 &10 &5 &10 &6 &10 &10 &10 &10 &10 \\
\# F &10 &10 &10 &10 &10 &10 &10 &10 &10 &10 &10 &10 \\
MD (h) &1.7 &1.5 &1.5 &1.5 &1.5 &1.6 &1.5 &1.5 &1.5 &1.5 &1.4 &1.5 \\
FD (h) &1.5 &1.5 &1.5 &1.5 &1.5 &1.5 &1.5 &1.5 &1.5 &1.7 &1.7 &1.6 \\
\midrule
\multicolumn{13}{@{}l}{\textbf{Noisy Valid}}  \\
\midrule
\# M &9 &10 &10 &10 &6 &10 &4 &10 &10 &10 &10 &10 \\
\# F &10 &10 &10 &10 &10 &10 &10 &10 &10 &10 &10 &10 \\
MD(h) &2.5 &2.5 &2.5 &2.0 &0.3 &1.8 &2.5 &0.9 &2.5 &2.5 &2.5 &2.5 \\
FD(h) &2.5 &2.5 &2.5 &2.5 &2.5 &2.5 &2.5 &1.7 &2.5 &2.5 &2.5 &2.5 \\
\bottomrule

\end{tabular}
\caption{Language wise distribution of audio data in hours for clean and noisy data. (\# M= Number of male speakers, \# F= Number of female speakers, MD(h) = Male audio duration, FD(h) = Female audio duration in hours)}
\label{tab:language_wise_distribution}
\end{table*}
\endgroup

\subsection{Speech Language Understanding Tasks}
We now described the different tasks covered in IndicSUPERB and the method used for creating training and evaluation data for these tasks.

\noindent\textbf{Automatic Speech Recognition (ASR).} ASR is the task of transcribing a given audio utterance into text. For training and evaluating an ASR system, we need an audio utterance paired with its text transcript. Since the collected data already contains this alignment by design, no extra processing was needed to create training or validation data for ASR. We simply use the aligned audio-transcript pairs in the splits described earlier for training, validation and testing.

\noindent\textbf{Language Identification (LID).} LID is the task of identifying the language of an audio clip. In other words, the task is to take the raw audio as input and classify it into one of the given $n$ languages ($n=12$ in this case). To create training data for this task, we simply pooled the audio recordings from all the languages in the ASR training data described earlier and assigned the known language id as the target label. We thus get labelled data for a $n$-class classification problem. We similarly create the multiple test and validation sets from the corresponding splits for ASR.

\noindent\textbf{Speaker Identification (SID).} SID is the task of identifying the speaker in an audio clip. In other words, the task is to take the raw audio as input and classify it into one of the given $k$ speaker IDs. We consider two scenarios here. First, where we try to identify speakers within a given language (\textit{i.e.}, all utterances and speakers belong to one language only). We refer to this as SID-mono. To create training data for this task, we simply pooled the audio recordings from all the speakers for a given language in the ASR training data described earlier and assigned the known speaker id as the target label. We thus get labelled training data for a $k$-class classification problem for each of the 12 languages (of course, the value of $k$ is different for different languages). We, similarly create the (noisy-)test-known and (noisy-)validation data for SID-mono from the corresponding splits for ASR. Note that since this is a $k$-class classification problem, there cannot be unknown speakers/classes at test time.

Next, we consider a multilingual setup wherein to create training data, we pool the audio recordings from all the speakers from all the languages in the ASR training data described earlier and assigned the known speaker id as the target label. We thus get labelled training data for a $m$-class classification problem (where $m$ is the sum of the number of speakers across all languages). This not only gives us a larger pool of speakers but also allows us to evaluate SID in a language-agnostic setting. We refer to this as SID-multi. We, similarly create the (noisy-)test-known, and (noisy-)validation data for SID-mono  from the corresponding splits for ASR.

\begingroup
\setlength{\tabcolsep}{4pt} 
\renewcommand{\arraystretch}{0.8} 
\begin{table*}[th!]
    \centering
    \small
    \begin{tabular}{@{}lrrrrrrrrrrrr@{}}
    \toprule
    \textbf{} &\textbf{bn} &\textbf{gu} &\textbf{hi} &\textbf{kn} &\textbf{ml} &\textbf{mr} &\textbf{or} &\textbf{pa} &\textbf{sa} &\textbf{ta} &\textbf{te} &\textbf{ur} \\
    \midrule
    \textbf{\#U }&839 &1046 &1049 &1046 &1049 &1031 &942 &732 &963 &1047 &1048 &1045 \\
\textbf{\#Q} &50 &50 &50 &50 &50 &50 &50 &50 &50 &50 &50 &50 \\
    \bottomrule
    \end{tabular}
    \caption{Number of utterances and queries in the data for QbE task. (\#U = Number of utterances, \#Q = Number of queries)}
    \label{tab:qbe_table}
    \end{table*}
\endgroup

\noindent\textbf{Automatic Speaker Verification (ASV).} In ASV, the input contains a pair of utterances and the task is to identify if the two utterances belong to the same speaker or not (the identity of the speaker does not matter). For this, we follow \cite{superb} and train a speaker identification model using the SID training data described above. We then compute speaker embeddings using this model and compute the cosine similarity between the embeddings of two utterances to decide if they belong to the same speaker. To create the test and validation sets for this binary classification problem we make 50000 pairs of utterances from the corresponding split of the ASR data described earlier such that 50\% of the pairs belong to the same speaker and 50\% of them belong to different speakers. Note that the transcripts  corresponding to the raw audio are used only for the ASR task and not for any of the other tasks.

\noindent\textbf{Query by Example (QbE).} In QbE, the task is to fetch all the audio clips which contain the spoken query word. Unlike text-based retrieval, here the query is also an audio file and the collection also contains audio files.  The ASR read speech data that we had collected contained only sentences and no queries. Hence, for this task, we did a separate data collection exercise where we first curated 50 popular entity names in each language (names of states, cities, celebrities, \textit{etc.}). For every query, we mined 20 sentences from publicly available sources, such as IndicCorp and Wikipedia resulting in a total of approximately 1K sentences for each language (please see Table \ref{tab:qbe_table} for the exact number of utterances and queries per language). We can keep/remove that based on the need. Once we had the queries and the corresponding sentence containing these query words, we set up a team of volunteers to read out the queries and the sentences. For every language, we involved a total of 20 volunteers, 10 for speaking out the queries and 10 for speaking out the sentences. There was an equal proportion of male and female speakers. Around 40\% of the speakers were from the 20-30 year age groups and 20\% each from the 15-20, 30-45 and 45+ age groups. Note that since this is a retrieval task there is no training involved here. In summary, the benchmark for this task contains 50 queries and a collection of 1000 utterances from which only 20 are relevant for a given query. 

\begingroup
\setlength{\tabcolsep}{4pt} 
\renewcommand{\arraystretch}{0.8} 
\begin{table*}[ht!]
\centering
\small
\begin{tabular}{@{}lcccccccccccc@{}}
\toprule
\textbf{} &\textbf{bn} &\textbf{gu} &\textbf{hi} &\textbf{kn} &\textbf{ml} &\textbf{mr} &\textbf{or} &\textbf{pa} &\textbf{ta} &\textbf{te} &\textbf{ur} \\
\midrule
\textbf{\#Tr} &2538 &2031 &2206 &8820 &6033 &1620 &1360 &8204 &1815 &569 &4054 \\
\textbf{\#Vl} &314 &245 &270 &1135 &783 &195 &155 &1051 &214 &54 &510 \\
\textbf{\#Ts}  &612 &492 &552 &2177 &1512 &382 &318 &2027 &439 &109 &989 \\
\midrule
\textbf{UPK} &69 &55 &61 &243 &167 &44 &37 &226 &49 &15 &111 \\
\bottomrule
\end{tabular}
\caption{Train, Valid and Test splits for KS task. \#Tr = Number of utterances in training set, \#Vl = Number of utterances in validation set, \#Ts = Number of utterances in the test set.}
\label{tab:ldcil_utterance_table}
\end{table*}
\endgroup

\noindent\textbf{Keyword Spotting (KS).} KS is the task of classifying an utterance into a predefined set of keywords. The keywords are usually commands such as up, down, open, close, \textit{etc}. We use the command and control words from the LDCIL dataset for all the 12 languages except Sanskrit. We took the top 50 frequent words in every language from the command and control category. These keywords were spoken by multiple speakers and we used these utterances to create the train, valid, test splits as shown in Table \ref{tab:ldcil_utterance_table}.

\begingroup
\begin{table*}[!t]
    \centering

    \small
    \begin{tabular}{@{}lccc|cccccc|c|c@{}}
    \toprule
    \textbf{Data}&\multicolumn{3}{c|}{\textbf{MSR}}&\multicolumn{6}{c|}{\textbf{MUCS}} & \textbf{OSLR} &  \\
    \textbf{} &\textbf{gu} &\textbf{ta} &\textbf{te} &\textbf{gu} &\textbf{hi} &\textbf{mr} &\textbf{or} &\textbf{ta} &\textbf{te} &\textbf{bn} &\textbf{avg} \\
    \midrule
All Existing &14.3 &\textbf{17.8} &14.1 &17.8 &12.9 &14.1 &\textbf{23.4} &19.5 &16.1 &11.7 &16.2 \\
\quad + \dataset &\textbf{11.2} &18.6 &\textbf{13.0} &\textbf{12.2} &\textbf{10.1} &\textbf{13.0} &24.3 &\textbf{19.1} &\textbf{13.4} &\textbf{11.5} &\textbf{14.7} \\
    \bottomrule
    
    \end{tabular}
    \caption{
    Comparisons of models trained using all existing data and existing data+\dataset~on MUCS, MSR and OpenSLR benchmarks.}
    \label{tab:tb2_bm_bmnew_compare}
\end{table*}
\endgroup

\begingroup
\renewcommand{\arraystretch}{1} 
\begin{table}[ht]
    \centering
    \small
    \begin{tabular}{@{}lcccc@{}}
    \toprule
    &\multicolumn{2}{c}{\textbf{Test Clean}} & \multicolumn{2}{c}{\textbf{Test Noisy}} \\
    \textbf{Model} & \textbf{Known} & \textbf{Unknown} & \textbf{Known} & \textbf{Unknown} \\
    \midrule
    \multicolumn{5}{@{}l}{\textbf{Language Identification (LID) - Acc.}} \\
    \midrule
    FBANK &27 &14.10 &26.15 &13.37 \\
    IndicW2V &\textbf{98.24} &\textbf{90.78} &\textbf{97.7} &\textbf{87.29} \\
    XLS-R &94.38 &79.96 &92.97 &74.50 \\
    \midrule
    \multicolumn{5}{@{}l}{\textbf{Speaker Identification Multilingual (SID-Multi) - Acc.}} \\
    \midrule
    FBANK &36.79 &- &36.23 &- \\
    IndicW2V &\textbf{79.26} & &\textbf{78.08} & \\
    XLS-R  &70.71 &- &69.22 & -\\
    \bottomrule
    \end{tabular}
    \caption{
    Comparison of different feature representations for the tasks of LID, SID-Multi. Note that for SID, we do not have any unknown speakers.}
    \label{tab:tb3_summ}
\end{table}
\endgroup
\begingroup

\begingroup
\begin{table*}[h]
\centering
\small
\begin{tabular}{@{}lrrrrrrrrrrrr|r@{}}
\toprule
\textbf{} &\textbf{bn} &\textbf{gu} &\textbf{hi} &\textbf{kn} &\textbf{ml} &\textbf{mr} &\textbf{or} &\textbf{pa} &\textbf{sa} &\textbf{ta} &\textbf{te} &\textbf{ur} &\textbf{avg} \\
\midrule
\multicolumn{13}{@{}l}{\textbf{Test Known}} \\
\midrule
FBANK &84.3 &68.6 &76.8 &83.2 &92.5 &68.5 &79.4 &76.0 &80.5 &72.5 &75.6 &68.3 &77.2 \\
IndicW2V &\textbf{99.3} &\textbf{93.2} &\textbf{97.4} &\textbf{98.1} &\textbf{99.8} &\textbf{86.4} &\textbf{98.0} &\textbf{96.3} &\textbf{98.6} &\textbf{90.6} &\textbf{92.3} &\textbf{97.3} &\textbf{95.6} \\
XLS-R &99.3 &93.0 &95.6 &97.7 &99.7 &83.1 &97.6 &92.5 &96.9 &87.7 &90.4 &97.3 &94.2 \\
\midrule
\multicolumn{13}{@{}l}{\textbf{Noisy Test Known}} \\
\midrule
FBANK &82.6 &64.0 &74.6 &85.4 &89.8 &65.4 &78.1 &78.6 &67.8 &63.1 &76.1 &80.1 &75.5 \\
IndicW2V &\textbf{99.3} &\textbf{90.9} &\textbf{93.0} &\textbf{98.7} &\textbf{99.6} &\textbf{89.3} &\textbf{97.8} &\textbf{96.7} &\textbf{96.5} &\textbf{88.8} &\textbf{93.7} &\textbf{97.8} &\textbf{95.2} \\
XLS-R &99.0 &89.8 &88.2 &98.7 &99.6 &82.5 &96.4 &94.0 &90.5 &81.8 &91.1 &97.2 &92.4 \\
\bottomrule
\end{tabular}
\caption{SID-Mono accuracy on Known split of Kathbath}
\label{tab:sid_mono}
\end{table*}
\endgroup

\begingroup
\setlength{\tabcolsep}{4pt} 
\renewcommand{\arraystretch}{0.8} 
\begin{table*}[h!]
\centering
\small
\begin{tabular}{@{}lrrrrrrrrrrrr|r@{}}
\toprule
\textbf{model} &\textbf{bn} &\textbf{gu} &\textbf{hi} &\textbf{kn} &\textbf{ml} &\textbf{mr} &\textbf{or} &\textbf{pa} &\textbf{sa} &\textbf{ta} &\textbf{te} &\textbf{ur} &\textbf{avg} \\
\midrule
\multicolumn{13}{@{}l}{\textbf{Test Known}} \\
\midrule
FBANK &2.1 &4.8 &3.4 &1.5 &0.7 &6.3 &3.4 &3.0 &0.4 &6.0 &2.9 &1.0 &3.0 \\
IndicW2V &2.0 &\textbf{3.3} &\textbf{2.6} &0.8 &0.4 &5.6 &\textbf{1.5} &\textbf{0.2} &\textbf{0.2} &\textbf{5.1} &\textbf{2.8} &\textbf{0.3} &\textbf{2.1} \\
XLS-R &\textbf{1.3} &3.9 &3.6 &\textbf{0.3} &\textbf{0.0} &\textbf{4.2} &2.2 &0.5 &0.3 &5.8 &3.4 &0.4 &2.1 \\
\midrule
\multicolumn{12}{@{}l}{\textbf{Test Unknown}} \\
\midrule
FBANK &13.3 &16.7 &9.0 &10.9 &\textbf{13.8} &8.8 &\textbf{11.0} &5.7 &8.3 &14.4 &11.6 &\textbf{11.4} &\textbf{11.2} \\
IndicW2V &16.6 &21.7 &8.8 &17.6 &24.4 &12.8 &21.2 &7.1 &9.5 &15.3 &15.4 &13.5 &15.3 \\
XLS-R &\textbf{10.7} &\textbf{16.4} &\textbf{6.3} &12.2 &21.8 &10.8 &14.9 &\textbf{4.5} &\textbf{6.8} &15.7 &11.9 &12.6 &12.1 \\
\midrule
\multicolumn{13}{@{}l}{\textbf{Noisy Test Known}} \\
\midrule
FBANK &2.2 &6.6 &5.5 &0.4 &1.2 &5.3 &8.5 &4.2 &1.0 &6.8 &3.6 &2.9 &4.0 \\
IndicW2V &2.8 &\textbf{3.1} &\textbf{2.9} &0.6 &0.1 &4.3 &\textbf{1.9} &\textbf{1.4} &\textbf{1.0} &\textbf{2.8} &\textbf{1.9} &\textbf{2.4} &\textbf{2.1} \\
XLS-R &\textbf{2.0} &3.5 &5.2 &\textbf{0.6} &\textbf{0.0} &\textbf{5.1} &4.0 &1.7 &0.8 &5.5 &2.5 &3.0 &2.8 \\
\midrule
\multicolumn{13}{@{}l}{\textbf{Noisy Test Unknown}} \\
\midrule
FBANK &12.8 &16.9 &10.0 &\textbf{10.2} &\textbf{12.4} &\textbf{8.6} &\textbf{10.5} &7.0 &8.2 &14.7 &12.4 &\textbf{11.3} &\textbf{11.2} \\
IndicW2V &15.7 &21.2 &9.4 &17.3 &21.5 &12.8 &22.3 &7.6 &9.8 &16.4 &14.7 &16.1 &15.4 \\
XLS-R &\textbf{11.9} &\textbf{16.4} &\textbf{6.8} &10.6 &16.3 &10.7 &14.3 &\textbf{5.5} &\textbf{7.6} &14.7 &12.7 &11.6 &11.6 \\

\bottomrule
\end{tabular}
\caption{ASV results on Kathbath using different models. (EER)}
\label{tab:asv_all}
\end{table*}
\endgroup

\begingroup
\setlength{\tabcolsep}{4pt} 
\renewcommand{\arraystretch}{0.8} 
\begin{table*}[ht]
    \centering
    \small
    \begin{tabular}{@{}lrrrrrrrrrrrr|r@{}}
    \toprule
\textbf{} &\textbf{bn} &\textbf{gu} &\textbf{hi} &\textbf{kn} &\textbf{ml} &\textbf{mr} &\textbf{or} &\textbf{pa} &\textbf{sa} &\textbf{ta} &\textbf{te} &\textbf{ur} &\textbf{avg} \\
    \midrule
    \multicolumn{14}{@{}l}{\textbf{KS (Accuracy)}}\\
    \midrule
    FBANK &15.0 &20.1 &18.5 &26.4 &19.4 &21.7 &15.7 &28.1 &-&20.5 &23.9 &27.0 &21.5 \\
IndicW2V &\textbf{97.2} &97.6 &\textbf{99.3} &\textbf{99.4} &95.4 &\textbf{97.4} &97.2 &98.2 &-&\textbf{98.9} &88.1 &97.2 &96.9 \\
XLS-R &96.9 &\textbf{98.8} &\textbf{99.3} &\textbf{99.4} &\textbf{96.1} &97.1 &97.8 &\textbf{98.6} &-&98.4 &\textbf{89.0} &\textbf{97.3} &97.1 \\

    \midrule
    \multicolumn{14}{@{}l}{\textbf{QbE (MTWV)}}\\
    \midrule
    FBANK &0.000 &0.001 &0.000 &0.000 &0.002 &0.000 &0.000 &0.000 &0.006 &0.000 &0.000 &0.000 &0.001 \\
IndicW2V &\textbf{0.026} &\textbf{0.003} &\textbf{0.004} &\textbf{0.023} &\textbf{0.017} &\textbf{0.046} &0.007 &0.008 &\textbf{0.071} &\textbf{0.021} &\textbf{0.012} &\textbf{0.023} &\textbf{0.022} \\
XLS-R &0.014 &0.001 &0.001 &0.023 &0.009 &0.015 &\textbf{0.010} &0.003 &0.040 &0.001 &0.011 &0.018 &0.012 \\

    \bottomrule
    \end{tabular}
    \caption{
    Comparison of different feature representations for the tasks of keyword spotting (accuracy) and query-by-example (MTWV).}
    \label{tab:tb4_ks_qbe}
\end{table*}
\endgroup

\begingroup
\renewcommand{\arraystretch}{0.8} 

\begingroup
\setlength{\tabcolsep}{4pt} 
\renewcommand{\arraystretch}{0.8} 
\begin{table*}[h!]
\centering
\small
\begin{tabular}{@{}lrrrrrrrrrrrr|r@{}}
\toprule
\textbf{} &\textbf{bn} &\textbf{gu} &\textbf{hi} &\textbf{kn} &\textbf{ml} &\textbf{mr} &\textbf{or} &\textbf{pa} &\textbf{sa} &\textbf{ta} &\textbf{te} &\textbf{ur} &\textbf{avg} \\
\midrule
Male &10.1 &8.4 &10.9 &8.4 &24.6 &9.5 &17.2 &7.0 &28.9 &7.1 &10.3 &11.5 &12.8 \\
Female &10.1 &5.6 &5.8 &9.3 &24.4 &7.7 &15.6 &5.5 &31.1 &16.7 &13.0 &14.5 &13.3 \\

\bottomrule
\end{tabular}
\caption{WERs results using IndicWav2Vec (with LM) for male and female sub-splits of test-unknown split across languages}
\label{tab:gender_bias}
\end{table*}
\endgroup

\begingroup
\setlength{\tabcolsep}{4pt} 
\renewcommand{\arraystretch}{0.8} 
\begin{table*}[ht!]
\centering
\small
\begin{tabular}{@{}lrrrrrrrrrrrr|r@{}}
\toprule
\textbf{} &\textbf{bn} &\textbf{gu} &\textbf{hi} &\textbf{kn} &\textbf{ml} &\textbf{mr} &\textbf{or} &\textbf{pa} &\textbf{sa} &\textbf{ta} &\textbf{te} &\textbf{ur} &\textbf{avg} \\
\midrule
\multicolumn{13}{@{}l}{\textbf{Test Known}} \\
\midrule
IndicW2V &13.3 &15.6 &10.7 &20.1 &31.4 &20.0 &19.5 &14.2 &30.8 &23.9 &24.5 &15.8 &20.0 \\
\quad + LM &\textbf{9.6} &\textbf{5.7} &\textbf{7.4} &\textbf{8.7} &23.0 &\textbf{8.4} &15.4 &\textbf{5.3} &\textbf{29.9} &13.0 &9.9 &\textbf{12.6} &\textbf{12.4} \\
XLS-R &14.8 &17.2 &12.1 &20.3 &32.2 &19.8 &20.3 &15.7 &33.9 &23.5 &24.1 &15.9 &20.8 \\
\quad + LM &9.9 &5.8 &7.5 &8.7 &\textbf{22.9} &8.5 &\textbf{15.2} &5.5 &30.1 &\textbf{12.8} &\textbf{9.6} &12.8 &12.5 \\
\midrule
\multicolumn{12}{@{}l}{\textbf{Test Unknown}} \\
\midrule
IndicW2V &14.3 &17.4 &12.2 &20.8 &37.8 &20.4 &22.2 &15.1 &37.0 &27.4 &26.9 &16.0 &22.3 \\
\quad + LM &\textbf{10.1} &\textbf{7.0} &\textbf{8.3} &\textbf{8.9} &\textbf{24.5} &\textbf{8.6} &\textbf{16.4} &\textbf{6.3} &\textbf{30.0} &12.5 &11.7 &\textbf{13.0} &\textbf{13.1} \\
XLS-R &16.8 &20.0 &14.4 &21.6 &40.5 &20.2 &23.3 &17.2 &42.2 &27.7 &26.7 &16.7 &23.9 \\
\quad + LM &10.7 &7.5 &8.9 &9.0 &24.7 &9.2 &16.6 &6.9 &31.0 &\textbf{12.1} &\textbf{11.6} &13.5 &13.5 \\
\midrule
\multicolumn{13}{@{}l}{\textbf{Noisy Test Known}} \\
\midrule
IndicW2V &15.3 &18.6 &13.2 &21.0 &33.2 &20.3 &22.3 &15.1 &41.7 &28.1 &27.2 &18.0 &22.8 \\
\quad + LM &\textbf{11.1} &\textbf{7.2} &\textbf{9.3} &9.1 &24.0 &9.3 &\textbf{17.5} &\textbf{6.7} &\textbf{34.5} &16.1 &12.5 &\textbf{14.4} &\textbf{14.3} \\
XLS-R &17.4 &20.3 &14.6 &20.2 &33.4 &19.9 &23.4 &15.9 &45.9 &27.6 &26.4 &18.4 &23.6 \\
\quad + LM &11.4 &7.5 &9.3 &\textbf{8.8} &\textbf{23.8} &\textbf{9.1} &17.6 &6.7 &35.6 &\textbf{15.7} &\textbf{11.6} &14.8 &14.3 \\
\midrule
\multicolumn{13}{@{}l}{\textbf{Noisy Test Unknown}} \\
\midrule

IndicW2V &18.2 &22.8 &14.1 &29.0 &40.7 &22.4 &25.3 &22.9 &45.5 &27.6 &30.6 &21.8 &26.7 \\
\quad + LM &\textbf{12.8} &\textbf{10.1} &\textbf{9.2} &13.6 &25.9 &10.1 &18.2 &\textbf{10.5} &\textbf{34.4} &15.0 &16.4 &\textbf{17.7} &\textbf{16.2} \\
XLS-R &21.1 &26.7 &16.6 &29.4 &43.5 &23.4 &27.4 &25.8 &50.5 &27.0 &29.8 &22.5 &28.6 \\
\quad + LM &13.1 &10.4 &9.7 &\textbf{13.5} &\textbf{25.9} &\textbf{10.8} &\textbf{18.1} &10.8 &35.4 &\textbf{14.8} &\textbf{14.4} &18.1 &16.2 \\
\bottomrule
\end{tabular}
\caption{WERs across different splits of Kathbath dataset using monolingual models trained on different SSL backbones. }
\label{tab:asr_monolingual}
\end{table*}
\endgroup

\begingroup
\setlength{\tabcolsep}{4pt} 
\renewcommand{\arraystretch}{0.8} 
\begin{table*}[h!]
\centering
\small
\begin{tabular}{@{}lrrrrrrrrrrrr|r@{}}
\toprule
\textbf{} &\textbf{bn} &\textbf{gu} &\textbf{hi} &\textbf{kn} &\textbf{ml} &\textbf{mr} &\textbf{or} &\textbf{pa} &\textbf{sa} &\textbf{ta} &\textbf{te} &\textbf{ur} &\textbf{avg} \\
\midrule
\multicolumn{13}{@{}l}{\textbf{Test Known}} \\
\midrule
IndicW2V &18.4 &18.6 &12.7 &24.1 &36.3 &23.0 &22.3 &17.4 &41.6 &29.1 &28.9 &21.5 &24.5 \\
\quad + LM &10.8 &6.0 &8.0 &9.2 &23.2 &9.5 &15.7 &5.9 &30.8 &14.0 &10.3 &13.6 &13.1 \\
XLS-R &19.4 &19.7 &14.3 &24.7 &36.9 &23.7 &23.2 &18.5 &39.1 &28.8 &28.6 &21.1 &24.8 \\
\quad + LM &\textbf{10.6} &\textbf{5.8} &\textbf{7.9} &\textbf{9.0} &\textbf{22.6} &\textbf{8.9} &\textbf{15.6} &\textbf{5.7} &\textbf{30.3} &\textbf{13.5} &\textbf{10.1} &\textbf{13.6} &\textbf{12.8} \\
\midrule
\multicolumn{12}{@{}l}{\textbf{Test Unknown}} \\
\midrule
IndicW2V &19.3 &20.3 &17.4 &24.1 &43.1 &22.8 &28.9 &19.0 &47.0 &33.2 &30.5 &41.6 &28.9 \\
\quad + LM &10.9 &7.2 &10.3 &9.3 &24.6 &9.4 &20.4 &7.3 &31.3 &13.4 &12.3 &25.5 &15.2 \\
XLS-R &20.8 &22.0 &19.6 &25.1 &43.6 &24.5 &27.2 &20.4 &44.9 &32.9 &30.8 &43.5 &29.6 \\
\quad + LM &\textbf{10.9} &\textbf{7.2} &\textbf{9.8} &\textbf{8.7} &\textbf{24.2} &\textbf{8.8} &\textbf{17.3} &\textbf{7.1} &\textbf{31.1} &\textbf{12.7} &\textbf{11.8} &\textbf{22.1} &\textbf{14.3} \\
\midrule
\multicolumn{13}{@{}l}{\textbf{Noisy Test Known}} \\
\midrule
IndicW2V &21.1 &22.1 &15.3 &25.4 &38.1 &23.4 &25.6 &18.4 &54.2 &33.0 &31.5 &23.7 &27.6 \\
\quad + LM &12.5 &7.7 &10.0 &9.8 &24.5 &10.1 &18.3 &7.5 &37.0 &17.2 &13.1 &15.7 &15.3 \\
XLS-R &22.4 &22.9 &16.7 &25.2 &37.9 &25.6 &26.4 &19.1 &51.1 &32.7 &31.6 &23.8 &28.0 \\
\quad + LM &\textbf{12.1} &\textbf{7.4} &\textbf{9.5} &\textbf{9.2} &\textbf{23.8} &\textbf{9.5} &\textbf{18.0} &\textbf{7.0} &\textbf{35.8} &\textbf{16.4} &\textbf{12.6} &\textbf{15.4} &\textbf{14.7} \\
\midrule
\multicolumn{13}{@{}l}{\textbf{Noisy Test Unknown}} \\
\midrule
IndicW2V &24.3 &27.1 &19.9 &32.2 &44.5 &24.9 &32.9 &27.6 &56.9 &32.3 &34.3 &49.6 &33.9 \\
\quad + LM &14.4 &11.3 &\textbf{11.8} &14.3 &25.6 &10.9 &22.6 &\textbf{12.2} &36.2 &15.8 &17.0 &30.9 &18.6 \\
XLS-R  &26.1 &29.1 &24.9 &32.6 &45.4 &27.4 &30.6 &31.0 &54.1 &33.3 &34.2 &52.1 &35.1 \\
\quad + LM &\textbf{13.8} &\textbf{11.0} &12.5 &\textbf{14.2} &\textbf{24.8} &\textbf{10.3} &\textbf{18.9} &12.3 &\textbf{34.8} &\textbf{15.5} &\textbf{15.0} &\textbf{28.6} &\textbf{17.6 }\\
\bottomrule
\end{tabular}
\caption{WERs across different splits of Kathbath dataset using jointly fine-tuned multilingual model trained on different SSL backbones}
\label{tab:asr_multi}
\end{table*}
\endgroup

\section{Experimental Setup}
Using \dataset~and IndicSUPERB, our goal is to evaluate whether (i) \dataset~helps in improving ASR for Indian languages on existing benchmarks (ii) SSL models are effective feature encoders for downstream speech understanding tasks and (iii) existing models are robust under different evaluation conditions (such as noise, gender imbalance, unknown speakers). We now describe the experimental setup that we used for each of the 6 tasks to answer the above questions.

\textbf{ASR}: Recent work on Indian languages has shown that wav2vec2 based models perform well on a wide variety of benchmarks \cite{javed2022towards}. Following this, we evaluate two wav2vec2 based models, \textit{viz.}, IndicWav2Vec and XLS-R \cite{xlsr}. The former is pre-trained using 17K hours of raw audio data across 40 Indian languages while the latter is trained on half a million hours of raw audio data from 128 languages including 11 of the 12 Indian languages considered in this work. We take the publicly available pretrained models and fine-tune them using the training split from \dataset. We use the CTC loss function \cite{ctc} and perform language specific fine-tuning, \textit{i.e.}, we build a separate acoustic model for each language instead of a joint multilingual model. We restricted the output vocabulary of the model to characters only. In addition to the acoustic model, we also trained a 6-gram KenLM language model for each language using all the sentences from IndicCorp (ranging from 8M to 87M for different languages). During decoding, we combine the scores of the acoustic model and the language model. We use a beam size of 128 and set the LM weight and word score parameter to 2 and -1 respectively. In addition, we also jointly fine-tune a model on Kathbath dataset. The jointly fine-tuned model is trained by pooling all the data across languages with output vocab as the union of the vocabulary across all languages. We call this joint model as multilingual. 
For all our experiments we use WER (word error rate) to measure the performance of different models.

\textbf{Other tasks:} For the remaining tasks, we compare the features extracted from log mel filterbanks (FBANK), with the representations from SSL models, \textit{viz.}, IndicWav2Vec and XLS-R. We use the s3prl\footnote{https://github.com/s3prl/s3prl} framework for training and evaluating the models for these tasks. The framework allows us to evaluate SSL models by extracting representation from different layers of the pretrained model.  For the three classification tasks, \textit{viz.},  \textbf{LID, SID(mono/multi), KS}, we take the extracted representations, mean pool them and train a linear classifier on top of them using the cross-entropy loss function. For all these tasks, we use accuracy as the evaluation metric.\\

For \textbf{ASV}, we use the same setup as \cite{superb} and train the X-Vector model on the extracted representations. We use the cosine similarity between the x-vectors to measure the similarity between a pair of utterances. We use equal error rate (EER) as the evaluation metric. EER is the location on a ROC curve where the false acceptance rate and false rejection rate are equal (the lower the EER the better)

For \textbf{QbE}, we run DTW\cite{giorgino2009computing} to compute the similarity score for each query document pair using the exponential cosine distance function as it was found to be the best distance function in \cite{superb}. We explore representations from all the layers of pretrained models on the valid set. We finally report the test numbers by taking the best layer based on the validation set. Following standard practice, we use maximum term weighted value (MTWV) as the evaluation metric which balances misses and false alarms. 

\section{Results and Discussions}

The results of our experiments are summarised in Tables \ref{tab:tb2_bm_bmnew_compare} to \ref{tab:gender_bias}. Based on these results, we try to answer the following questions:

\noindent\textbf{Does \dataset~ help in improving ASR for Indian languages?}
To assess the usefulness of \dataset, we compare the performance of two IndicWav2Vec models. The first model is trained on all existing data from publicly available benchmarks, \textit{i.e.}, MUCS\cite{diwan2021multilingual}, MSR\cite{srivastava18_sltu} and a subset of OpenSLR\cite{nebesi} obtained from the authors of \citet{9414961}. The training data contained 40 hours each for Gujarati, Tamil and Telugu and 95 hours each for Hindi, Marathi and Odia. The second model is trained on \dataset~in addition to the above existing data. We use the same 6-gram KenLM language model to decode the emissions for both the acoustic models. As is evident from the results in Table \ref{tab:tb2_bm_bmnew_compare}, adding \dataset~ reduces the WER for most of the languages with an average improvement of 1.5\%. Note that this is despite the fact that the data distribution in \dataset~ is very different from the distribution of the data in these benchmarks. For example, while \dataset~contains formally written sentences from the News domain, the MUCS/MSR benchmarks contain code-mixed sentences. 

\noindent\textbf{How effective are SSL models as feature encoders for Indian languages?} We refer to Tables \ref{tab:tb3_summ},\ref{tab:sid_mono} and \ref{tab:asv_all} for a comparison of different feature representations for the 3 classification tasks, \textit{viz.}, LID, SID and ASV. We observe that across all the evaluation conditions, the features extracted from IndicWav2Vec and XLS-R perform better than FBANK features. The only exception to this rule is the performance on the Test-Unknown and Noisy-Test-Unknown splits for the ASV task. Similarly, referring to Table \ref{tab:tb4_ks_qbe} for QBE and KS, we observe that FBANK based features are easily outperformed by the features extracted from the two SSL models. We thus conclude that similar to English, features extracted from SSL models are useful for speech understanding tasks for Indian languages. 

\noindent\textbf{Are language-family specific SSL models better than those pre-trained on a larger set of languages?}
We now compare the features extracted from IndicWav2Vec, which is pretrained using data for 40 Indian languages with XLS-R which is pretrained on a larger set of 128 languages. Referring to Tables \ref{tab:tb3_summ},\ref{tab:sid_mono} and \ref{tab:tb4_ks_qbe}, we observe that for LID, SID and QbE, features extracted from IndicWav2Vec perform better than those from XLS-R. For ASV, XLS-R performs better than IndicWav2Vec on the splits containing unknown speakers. This is intuitive as XLS-R is trained on a much larger and diverse set of speakers, which perhaps helps it to generalise better to unknown speakers. Lastly, for the task of QbE there is only a small difference between the performance of the features extracted from the two models.  

\noindent\textbf{How does the performance generalise to unknown speakers?}
Once again referring to Tables \ref{tab:tb3_summ} and \ref{tab:asv_all}, we observe that the performance for LID and ASV drops when the test splits contain unknown speakers. The drop in performance is significant for the task of ASV with an increase in EER to double digit numbers from single digit numbers for all the models with features from the two SSL models performing worse than FBANK features. Also, as mentioned earlier, the XLS-R model which is trained on a much larger set of speakers performs better than the IndicWav2Vec model for the ASV task. Similarly, Table \ref{tab:asr_monolingual} shows the performance of IndicWav2Vec fine-tuned on \dataset~for the task of ASR. Once again, we see a drop in performance on the test split containing unknown speakers as compared to the split containing known speakers. While the difference is smaller for ASR, these results do suggest that while creating SLU benchmarks it is important to have test splits containing novel speakers which are not seen during training.

\noindent\textbf{How robust are existing models to noise in the utterances?}
Referring to Tables \ref{tab:tb3_summ} to \ref{tab:asr_monolingual}, we see a slight drop in the performance for all the tasks when evaluating on the noisy splits as compared to the corresponding clean splits. As expected, the gap between the most favourable condition (\textit{i.e.}, clean test sets with known speakers) and the most unfavourable condition (\textit{i.e.}, noisy test sets with unknown speakers) is quite high for most tasks. We thus believe that these multiple test conditions in IndicSUPERB will allow a more robust evaluation of speech models for Indian languages.  

\noindent\textbf{Monolingual vs Multilingual setting?}
Table \ref{tab:asr_monolingual} and \ref{tab:asr_multi} refer to WERs obtained using monolingual models and jointly trained multilingual models trained on Kathbath dataset. We observe from Table \ref{tab:asr_multi} that in multilingual setting XLS-R performs slightly better than IndicWav2Vec with less than 1\% WER improvement across clean and noisy splits of Kathbath. In addition, Table \ref{tab:asr_monolingual} and \ref{tab:asr_multi} show that the performance difference between monolingual models and their multilingual counterpart is not significant, which makes our multilingual model a good candidate for deployment.

\noindent\textbf{Are existing models robust to gender bias in the training data?}
As is evident from Table \ref{tab:language_wise_distribution}, our test and validation sets are well balanced, i.e., they have an equal number of male and female speakers as well as an equal number of hours from male and female speakers. However, while collecting data at scale it is hard to maintain this balance across languages. For example, for some languages, we were able to find more female speakers easily despite the incentives being similar for male and female speakers. Similarly, for some languages, we were able to find fewer participants who did bulk of the work as most participants were not willing to work for a shorter duration. As a result, our training data has imbalances wherein for most languages we have more female speakers and/or more number of hours for female speakers. Using our balanced test sets, we can evaluate the effect of this imbalance in the training data. To do so, we consider the task of ASR and focus on 3 languages: (i) Gujarati which has fewer female speakers than male speakers (25 v/s 34)  but the total duration for female speakers is 3 times more than that for male speakers (88 v/s 28) (ii) Hindi which has an almost equal number of male and female speakers (48 v/s 53) but the total duration for female speakers is more than twice that for male speakers (95 v/s 42) and (iii) Kannada which has much fewer female speakers than male speakers (16 v/s 43) but the total duration for female speakers is almost twice that for male speakers (97 v/s 56). In Table \ref{tab:gender_bias} we separately present the results for the male and female speakers in the \textit{test-unknown} set. We observe that for Hindi and Gujarati where the number of female speakers as well as the duration of female speakers is higher the performance is clearly better for female speakers. For Kannada, even though the duration is higher for females, the number of female speakers is much smaller and as a result, there is a slight dip in the performance for female speakers as compared to male speakers. We also see a similar trend for Punjabi and Tamil languages. Any data collection exercise will always have such biases due to easier availability of data from certain sections of the population and hence it is important to create balanced evaluation sets, as in IndicSUPERB, to evaluate the bias in models trained on such datasets. 

\section{Conclusion}
In this work, we first present \dataset, a large scale ASR dataset for 12 Indian languages. Using the meta information in \dataset, we create a robust benchmark called IndicSUPERB for ASR, language identification, speaker identification and speaker verification containing different test conditions with known speakers, unknown speakers and noisy utterances. We also create a benchmark for query by example (a retrieval task) and keyword spotting. Through our experiments, we first show that the ASR training data in \dataset~helps in improving the performance on existing ASR benchmarks across languages even though the data in \dataset~has a different distribution than the existing test sets. 
Next, we use IndicSUPERB, to evaluate the efficacy of existing SSL models and show that they serve as good feature encoders for a variety of SSL tasks. Lastly, we show that the different test conditions in IndicSUPERB are useful for evaluating (i) the capability of existing models to generalise to unknown speakers (ii) the robustness of existing models to noise and (iii) the robustness of existing models to bias in training data.

\section*{Acknowledgements}
We would like to thank the Ministry of Electronics and Information Technology (MeitY\footnote{https://www.meity.gov.in/}) of the Government of India and the Centre for Development of Advanced Computing (C-DAC\footnote{https://www.cdac.in/index.aspx?id=pune}), Pune for generously supporting this work and providing us access to multiple GPU nodes on the Param Siddhi Supercomputer. We would like to thank the EkStep Foundation and Nilekani Philanthropies for their generous grant which went into hiring human resources as well as cloud resources needed for this work. We would like to thank DesiCrew for connecting us to native speakers for collecting data. We would like to thank Vivek Seshadri from Karya Inc. for helping setup the data collection infrastructure on the Karya platform. We would like to thank all the members of AI4Bharat team in helping create the Query by Example dataset. 

\bibliography{main}
\bibliographystyle{acl}




\end{document}